\documentclass[cameraready]{Interspeech}

\title{Towards participatory speech dataset curation: \\ A queer case study and conceptual framework}

\author[affiliation={1,2}]{Brooklyn}{Sheppard}
\author[affiliation={2}]{Anaelia}{Ovalle}
\author[affiliation={2}]{Adina}{Williams}
\author[affiliation={2}]{Levent}{Sagun}

\address{
    $^1$ University of Calgary \\
    $^2$ Meta FAIR 
}

\email{brooklyn.sheppard1@ucalgary.ca}


\usepackage{comment}
\usepackage{subcaption}

\begin{document}

\maketitle

\begin{abstract}
  In this paper, we motivate the need for a participatory speech dataset creation framework through a case study of the LGBTQIA+, or \textit{queer}, community - a community with documented concerns about AI and reported harms, including attempts to develop `gaydar' technologies that purportedly identify individuals as queer. We review common speech data collection practices, why these methods may be unsuitable for engaging with queer speakers, and discuss previous efforts in participatory AI with queer community engagement, as well as participatory endeavours specific to speech data collection for other marginalized communities. From this review, we develop a conceptual framework for participatory speech data curation by, for, and with marginalized communities drawing on insights from co-design and knowledge sharing. We propose a framework comprising overlapping and two-way processes of defining a community, project formulation, modes of participation, and personal autonomy.
\end{abstract}

\section{Introduction}
\label{sec:intro}

It is notoriously difficult to collect speech data that is both high-quality and diverse. In addition to simply recording the speech itself, a significant amount of labour from those within the language community is required to accurately annotate the data \cite{papakyriakopoulos2023augmented}, and further care should be taken when dealing with potentially sensitive annotations related to speaker identity. For example, while gender annotations are quite common across speech datasets, for those who do not identify within the gender binary, this information may identify them as being a member of the LGBTQIA+, or \textit{queer}, community.

Crowdsourcing is commonly used as an efficient method of speech data collection and annotation. Crowdsourcing efforts in speech data collection have resulted in datasets for specific languages \cite{abraham2020crowdsourcing}, specific dialects of a given language \cite{sanabria23edacc}, and specific demographic groups within a language community \cite{coraal}. While large-scale crowdsourcing is efficient, a gap exists when these efforts lack meaningful, long-term collaboration. This has led to concerns about so-called `participation washing,' where engagement is temporary rather than a sustained commitment \cite{sloane2022participation}. In contrast, participatory efforts in AI reflect the process of engaging with diverse stakeholders, bringing users, affected communities, researchers, and engineers together to collaboratively design AI models by, for, and with the affected community.

Recent advances in participatory AI argue against the approach of simply scaling up data collection efforts to be more inclusive of diverse perspectives. For example, there are many diverse speech datasets that aim to represent a variety of speaker backgrounds, including the Fairspeech and Casual Conversations datasets, however, despite including speaker gender labels, the majority of these datasets do not include any speakers outside the gender binary, and those that do only contain a small number of speakers from this group \cite{veliche2024towards, porgali2023casual}. In contrast, efforts that focus on a specific community, such as the Mid-Atlantic Gender Expansive Speech (MAGES) corpus for gender-diverse representation in speech, instead provide data for specific use-cases such as diversifying synthetic voices or further understanding the role of one's speech in portraying their queer identity \cite{szekely2024inclusive, hope2023nonbinary}. 

With participatory approaches gaining traction in responsible and ethical AI, we propose that it is beneficial to consider participation at all stages, including and as early as dataset creation, and that such methods are particularly well-suited for diversifying speech technology datasets. The main contributions of the present work are 1) to motivate the need for a participatory framework for speech dataset curation through the lens of queer speakers, and 2) to propose a conceptual framework for speech dataset curation by, for, and with marginalized communities; thus providing speech researchers with an accessible framework to engage in participatory approaches to speech data curation.

\section{Challenges in speech data diversity: The case of queer speakers}

We begin by describing the queer community, why this community is an important case study for inclusive speech technology, and the unique challenges associated with direct engagement between the queer and speech technology communities. 

\subsection{The queer community \& their voice(s)}

In this work, we describe the queer community as anyone who self-identifies as `possessing non-normative sexual identity, gender identity, and/or sexual characteristics` \cite{tomasev2021fairness}. The queer community serves as an intriguing case study for speech data collection for a variety of reasons. First, despite speaker gender being a common annotation in most speech datasets, gender diverse speakers have the added complexity that volunteering their gender identity essentially ``outs" them as being a part of this marginalized community in a way that is distinct from cisgender binary identifying individuals. Thus, while gender annotations are not commonly assumed to be sensitive annotations compared to demographic factors such as sexual orientation or disability status, they become inherently tied to sensitive category status when including diverse gender perspectives. Second, the queer community represents a group of speakers whose speech characteristics may change considerably through time and context, either as a result of voice coaching or hormone therapy, or when communicating with speakers within versus outside this community \cite{hughes2024acoustic, hamdan2018effect}. 

The inclusion of queer speech in datasets for speech technology research would be beneficial in various domains such as speech recognition, speech categorization, and speech synthesis. In terms of speech synthesis, many current modelling efforts aim to provide a diverse set of voices either as a part of voice cloning efforts for those who may soon or have already lost their ability to speak \cite{szekely2024inclusive, ogun20241000, snyder2023busting}, or to diversify the voice options for end-users of a system. Indeed, recent work on the perception of synthetic voices shows that when provided with the option to customize voices in AI-voice assistants, users are more likely to have a positive experience and increased trust in their AI assistant, and this increase in trust can go so far as to mitigate misinformation they had previously been exposed to. \cite{snyder2023busting}. 

In the realm of speech recognition or speech categorization (i.e., speaker verification, emotion classification, etc.), previous research has found that these models are biased along several demographic axes including gender, age, race and disability status \cite{slaughter2023pre}. In terms of gender biases, previous investigations found that ASR models perform significantly worse on female voices when they are underrepresented in the training data \cite{garnerin2019gender}, but due to lack of data, most of these investigations are limited to biases along the gender binary \cite{sanchez2024beyond}. One previous study did attempt to investigate potential performance disparities across three speaker gender categories - ``male", ``female", and ``other" from the Common Voice 16.0 dataset \cite{attanasio-etal-2024-twists}. These researchers found that all speech models tested performed worse for speakers of a particular gender, however, models differed in whether they favoured the ``male" for ``female" identifying speakers. Intriguingly, however, the authors note that the ``other" gender category was never favoured across any of the models tested. While these results do indeed suggest performance biases against speakers that do not identify within the gender binary, with a sample size of only 8 to 86 speakers depending on the language, the authors note that this investigation is merely preliminary, and that more data is needed to make generalizable conclusions. With such limited data, it is then difficult to advance research in speech technology that is appropriate and useful to queer users, as well as evaluate the performance of existing models on this group of speakers. In the following section, we discuss unique challenges of queer speech data collection through previous findings of AI harms to this community and common data collection practices.

\subsection{The queer community and AI}

The queer community is a salient case study for ethical speech data collection given the long history of AI and technological harms associated with this community. \cite{mcara2024ai} surveys the impacts of AI hype on the queer community noting several examples of how these tools have been used to harm the queer community, from claims of an AI ``gaydar" to using geolocation data to target trans healthcare facilities  for harm \cite{mcara2024ai}. 
Prior work reports disproportionate flagging of queer dialects as toxic when using AI models for content-moderation \cite{thiago2021fighting, haimson2021disproportionate}. Additionally, qualitative interviews with trans and non-binary users of voice AI assistants, such as Siri and/or Alexa, found that these users conveyed a deep lack of trust in AI developers, specifically mentioning concerns of their privacy, describing discomfort at the thought of being visible to AI developers, and overall doubts of the ethics of AI practices \cite{rincon2021speaking}. In addition to AI harms on the queer community, the data collection practices common in speech technology are also not typically designed to take queer perspectives into account. For example, gender annotation practices have historically been annotated based on perceived gender and limited to binary gender labels, which risks misgendering and further erasure of the queer community, although self-reported gender identity is becoming increasingly more common in speech datasets \cite{sanabria23edacc, porgali2023casual}. Thus, the queer community serves as a challenging case study for participatory design in speech data collection given the longstanding lack of trust between this community and AI developers, and the lack of queer representation in current crowdsourced speech datasets.

\section{Previous efforts in participatory AI}

Some recent efforts in participatory AI have focused on engaging directly with the queer community, particularly in the field of AI-assisted mental health resources \cite{liem2024reclaiming, joyce2024defining}. For example, the Participatory Queer AI Research in Mental Health (PARQAIR-MH) proposes to use a method which involves several rounds of anonymous surveys consulting queer community members in an effort to iteratively come to a consensus on desiderata for the design of mental health AI tools \cite{joyce2024defining}. This method allows for the development of a toolkit for AI researchers and health policy institutions for taking queer preferences into account during the design and development of mental health AI tools. The process is participatory in that the queer community is explicitly sought after for guidance on how best to create technologies that serve this community, however, the extent to which the queer experts can influence model design or data collection down the road is limited. 

When working with speech data, however, the imperative for community participation becomes even more pronounced. Members of a given community not only provide the speech recordings and associated annotations, but also bear the risks associated with voice data, including the potential for speaker identification. These risks are heightened in contexts where voice data could inadvertently disclose an individual’s membership in a marginalized community, thereby increasing the likelihood of harm. Previous work on sensitive speech data in the realm of speech related to neurological disorders proposes several desiderata for ethically curating such speech datasets, including informed consent, stringent security and privacy measures, and being transparent about which communities are represented and which are not \cite{mancini2025promoting}. 

A participatory approach to speech data curation is well-suited to mitigate the inherent challenges of curating speech datasets by centering the desires of the speech community at all stages of the design process \cite{li2024want}. For example, in contrast to the stringent security measures called for by \cite{mancini2025promoting}, the StammerTalk initiative \cite{li2024want} instead demonstrates a case where community members advocated for the public release of their co-designed dataset for stuttered speech. This initiative, initially created by speech AI researchers who identify as part of the stuttering community, engaged with the online stuttering community, inviting anyone who is interested and self-identifies as a stutterer to participate, with no restrictions on age, gender, or other demographic factors. Additionally, a post-data collection survey demonstrated that the process for both the researchers and the speakers contributed to feelings of empowerment through helping provide these resources to their community. The StammerTalk initiative is an excellent example of participatory speech dataset creation, however, there is still a distinction between data collectors and data contributors in that the core research team formulated the project design independently from the data contributors, or speakers, themselves. The question then, is how to value the diverse perspectives of both the researchers involved and the speakers themselves at all stages in the data curation process.

Over the past several decades the field of linguistics has also attempted to adopt more reciprocal and participatory methods for speech and language data collection from marginalized communities \cite{yamada2007collaborative, akumbu2024decolonization}. This is most often seen in field linguistics, where typically white, European descendant scientists have historically engaged in a one-way relationship in which the linguist extracts data from a speaker in an effort to further the scientific discipline rather than to benefit the language community in any way. More recently, participatory efforts in field linguistics have called for a more cyclical approach, where language speakers and the linguists they work with engage in knowledge transfer in both directions, often with the goal of revitalizing the language through documentation and curriculum development \cite{benedicto2007model}. This method of engaging with the community in a partnership of knowledge sharing allows the language community to be fully involved in the process from project goals to the evaluation of outcomes. Drawing on the previous work in participatory design and queer engagement in these various fields, we propose a framework for participatory speech dataset creation by, for, and with the queer community. 

\section{Conceptual framework for community-led speech data curation}

\begin{figure}[t!]
    \centering
    \begin{subfigure}{0.23\textwidth}        \includegraphics[width=1\linewidth]{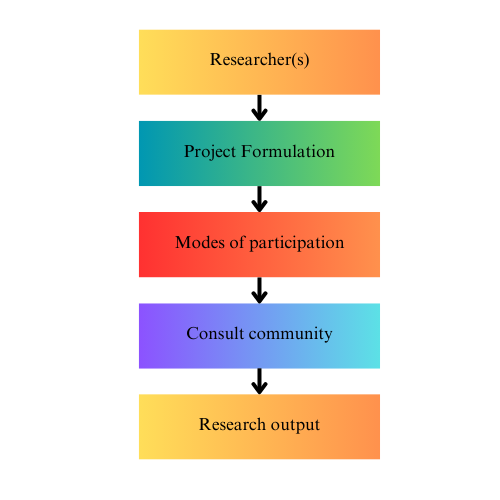}
    \caption{Top-Down Approach}
    \label{fig:ccv2}
    \end{subfigure}
    \begin{subfigure}{0.23\textwidth}
        \includegraphics[width=1\linewidth]{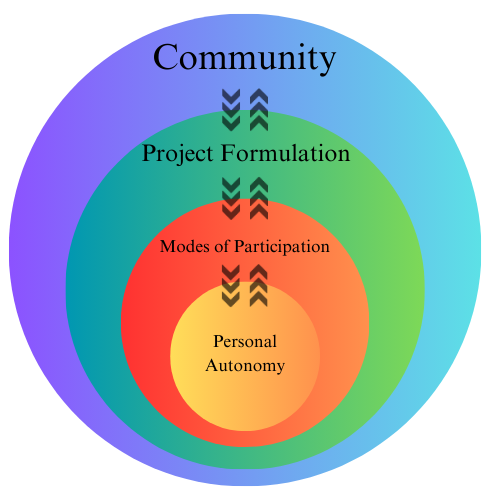}
    \caption{Co-Design Approach}
    \label{fig:edacc}
    \end{subfigure}
    \caption{Schematics of speech dataset creation frameworks. Left Panel: A one-directional flow from researcher to output; Right Panel: Nested circles with bidirectional arrows indicating iteration, collaboration, and co-design.}
    \label{fig1}
\end{figure}

The previous participatory efforts in engaging with the queer community and those focused on speech data collection offer several key insights into possible avenues for participatory speech dataset creation with the queer community. We summarize the comparison of traditional speech dataset collection efforts (e.g., crowdsourcing) to the proposed framework in Figure \ref{fig1}. First, central to all participatory work is the value of the community`s needs and lived experiences.  While previous efforts may differ in terms of whether they merely consult the affected community or fully co-design the project with the affected community, we highlight the utility of co-design when feasible for speech data curation given the fact that the community members themselves must be involved from recording the speech to annotating it, as well as the ones affected by this data being available for speech technology research. Second, in true co-design fashion, we envision the project formulation being encapsulated within the community itself, drawing on experiences from a diverse set of community members to inform the project goals and scope. The formulation of project goals and design then lead us to different possible modes of participation, such as contributing one`s voice, providing annotations of speech, or other labour related to project organization and documentation. Of course, the specific modes of participation may be different for each individual contributor, finally leading us to the level of personal autonomy over one`s own contributions and choices in terms of how their data may be used, how they would like their efforts to be acknowledged, and their personal level of involvement in the project. Lastly, the insights from linguistic field methods and co-design in terms of knowledge sharing suggest a two-way approach to honour the contributions of all parties involved. 

We present this framework as a form of normative infrastructure for speech dataset governance, rather than prescriptive set of requirements. We conceptualize this process through four phases of engagement that may relate to each other in a cyclical fashion. These phases are 1) community, 2) project formulation, 3) modes of participation, and 4) personal autonomy. Along these four phases, we provide questions to guide community contributors through the speech data curation process. 

\subsection{Community}

While many participatory approaches in AI begin with project conception or formulation, we instead focus first on the community the project is aimed to serve. That is, in true co-design fashion, the specifics of the project`s goals would benefit from input from the community members themselves. Thus, we begin our proposed engagement framework by first defining aspects related to the community members involved.

\textit{Who is this project going to serve?} In this case study, we take the term \textit{queer} to mean anyone who self-identifies as being a part of this community, including but not limited to queer researchers in speech technology and linguistics. 

\textit{Who might be left out and why?} Beginning with a broad approach to queer speech, we then consider who might be left out from this process. This includes varying intersections of identities within the queer community itself, such as race, disability status, and language or dialect. For example, if we target Queer in AI \cite{queerinai2023queer} -- a community of queer AI researchers -- this network is primarily composed of individuals from countries of the Global North and predominantly English speaking, thus potentially skewing the representation to be those with historical privilege. Different community engagement techniques, such as broadening the outreach to more queer community-based initiatives around the world may help alleviate this issue of representation. That being said, it is unlikely that all perspectives will be represented in a single project and thus members should be explicit about who is not represented. 

\subsection{Project formulation}

Once a general community is developed, the community members can work together to define the project scope more concretely in terms of genre(s) of speech and how the data will be used, stored and maintained over time.

\textit{What type of speech are we hoping to collect?} Given that speech of varying discourse genres may differ considerably (e.g., conversational speech versus read speech), it is important to be explicit about what genre of speech will be collected and consider the uses and risks of each. In the case of StammerTalk, for example, the researchers collected a variety of genres including both conversational speech discussing lived experiences with being a part of the stuttering community and a set of ASR commands \cite{li2024want}. In the case of the queer community, a similar approach may be helpful, but comes with its own challenges and risks. A conversational interview regarding queer perspectives and experiences could be ideal for assessing not only speech models on their accuracies of this type of speech within this community, but could also be helpful in identifying the relationship between queer dialects and higher toxicity ratings found in text models \cite{thiago2021fighting, haimson2021disproportionate}, and perhaps even mitigated by the presence of speech \cite{bell2024role}. That being said, this method risks the recording of personal and intimate conversations that may either dissuade participation or require a significant amount of data to ultimately be removed. 

\textit{Will the data be publicly available? What are the allowable uses of the data?} Given the risks associated with publicly available queer speech data, this is an important question for the queer community. That is, publicly available speech data may allow the queer community to be further targeted for harm, while in contrast, publicizing the data may contribute to the empowerment of community members to be out and proud. It is also deeply connected to the personal autonomy of each contributing speaker in whether they will allow their own voice to be a part of this dataset and how they will allow their personal contributions of data to be accessed or updated, such as in the case of fluctuating gender identities.

\textit{How will the data be stored? How will the data be maintained over time?} Given the issue of data sensitivity in the case of the queer community, it is important to be explicit about how the data could be stored and maintained over time. For example, how may community members revoke their data if they so choose? Will all members have access to the data indefinitely? The Databrary project is a particularly useful example of sensitive data sharing in the case of child behavioural video data which requires researchers hoping to access the data to have undergone ethics training and be affiliated with an institution that is governed by an ethics review board \cite{simon2015databrary}.

\subsection{Modes of participation}

Having defined the project scope, one can now turn to how participation might occur. In the realm of speech data, this includes how members may collaborate and how these contributions are acknowledged.

\textit{How will members collaborate together?} This pertains to not only the practical aspect of communication platforms and preferences among community members, but also about how community members agree to share the space and navigate power dynamics between the varying group members.  

\textit{What are the different ways collaborators can participate?} In developing a speech dataset, there are many forms of participation that one might choose to engage in. That is, some members may choose to provide voice recordings or annotations, while others may choose to take on an organizational role. 

\textit{How will the contributions of collaborators be acknowledged?} A common method of acknowledging contributions in speech data collection is of course where speakers are paid for their work \cite{li2024want, porgali2023casual}. Additionally, in field linguistics and in some more recent speech datasets for AI, authorship on any associated papers of the project may additionally be a desirable outcome for some \cite{netzorg2024speech}. How one would like their contributions to be acknowledged may be a highly individual preference, thus bringing us to the phase of personal autonomy.

\subsection{Personal autonomy}

Finally, we come to the issue of personal autonomy; that is, each individual`s right to engage in the process that is most empowering to them. This of course interacts with all other levels of the framework as not only is a community and its needs defined through a mosaic of the needs of individuals within that community, but the individual actions of each member may have downstream effects on the community as a whole.

\textit{At which points of the process do we expect group consensus versus individual choices?} The delicate balance between personal autonomy and community responsibility calls for careful consideration of this question. That is, in some cases the community may need to come to consensus on certain decisions, such as which genres of speech will be targeted or the overall goals of the project, while there is also ample room for community members to take their own personal stance specifically in terms of their modes of participation, how that participation is acknowledged, and the freedom to revoke their own data. That being said, this question of when to expect consensus is crucial when considering the potential downstream effects of any one individual's participation. 

\subsection{A two-way process}

Although we break down this process into four phases, we suggest that this process need not be unidirectional, and instead is necessarily a two-way process. While the process may begin with a given community, further specifications in the project formulation phase may warrant the need to reconsider who the community in question is. For example, if we begin with the queer community very generally, the question then arises of, for example, which languages and dialects within the queer community will be represented. This may develop differing directions in the project formulation stage which in turn affects which community or communities should be involved in the process. This is also the case in terms of the relationship between project formulation and personal autonomy. For example, for speakers who are at risk of being harmed based on identifying publicly as a member of the queer community, they may need to consider whether or not their data or contributions can be shared publicly or if their data needs to have more stringent security measures, if shared at all. Thus, this framework is a  dynamic process of continuously considering the levels of community, project formulation, modes of participation, and personal autonomy throughout the process of dataset creation.

\section{Conclusion and future directions}

The present paper provides AI researchers and queer community members with guidelines on the process of participatory speech dataset creation, through a collaborative two-way knowledge sharing approach. We take the queer community as a case study, but we believe the proposed framework can be applied to a variety of participatory speech dataset development projects outside of just the queer community. Future work would benefit from iteratively refining this framework for a variety of different use cases and speech communities. For example, if the proposed framework is used to develop a dataset of queer speech, further improvements could be made to the framework through reflections on the process itself and feedback from community members. We hope the present framework inspires more queer perspectives to be considered in a collaborative way for the advancement of speech technology development and research by, for, and with the queer community.

\section{Generative AI Use Disclosure}
The authors did not use any generative AI tools in any step of this research.

\bibliographystyle{IEEEtran}
\bibliography{mybib}

\end{document}